\documentclass{article}

\PassOptionsToPackage{numbers, compress}{natbib}

\usepackage[preprint]{neurips_2024}

\usepackage[utf8]{inputenc} 
\usepackage[T1]{fontenc}    
\usepackage{hyperref}       
\usepackage{url}            
\usepackage{booktabs}       
\usepackage{amsfonts}       
\usepackage{nicefrac}       
\usepackage{microtype}      
\usepackage{xcolor}         
\usepackage{graphicx}

\newcommand{\methodname}{{\tt{FedMRL}}}

\newcommand{\hetero}{heterogeneous }
\newcommand{\homo}{homogeneous }
\newcommand{\pers}{personalized }
\newcommand{\gen}{generalized }
\newcommand{\heteroN}{heterogeneity }

\newcommand{\persN}{personalization }

\newcommand{\sota}{state-of-the-art }
\newcommand{\rep}{representation }
\newcommand{\reps}{representations }
\newcommand{\assum}{Assumption }

\usepackage{amsmath}
\usepackage[ruled]{algorithm2e}
\usepackage{tcolorbox}
\definecolor{table_color1}{HTML}{EFEFEF}
\definecolor{table_color2}{HTML}{9B9B9B}
\definecolor{table_color3}{HTML}{C0C0C0}
\definecolor{table_edge}{RGB}{0,0,0}
\newtcbox{\mybox}[1][red]{on line, colback = {RGB}{255,228,225}, colframe = {RGB}{255,193,193},  arc=1mm, auto outer arc, boxrule=0.5pt,}
\usepackage{multirow}
\usepackage{colortbl}

\newtheorem{theorem}{Theorem}
\newtheorem{lemma}{Lemma}
\newtheorem{proof}{Proof}
\newtheorem{assumption}{Assumption}

\usepackage{enumitem} 

\newlist{circledenum}{enumerate}{1}
\setlist[circledenum,1]{label=\textcircled{\arabic*}}

\title{Federated Model Heterogeneous \\ Matryoshka Representation Learning}

\author{%
Liping Yi$^{1,2}$, Han Yu$^{2}$, Chao Ren$^{2}$, Gang Wang$^{1}$, Xiaoguang Liu$^{1}$, Xiaoxiao Li$^{3}$ \\
$^1$College of Computer Science, TMCC, SysNet, DISSec, GTIISC, Nankai University\\
$^2$College of Computing and Data Science, Nanyang Technological University\\
$^3$Department of Electrical and Computer Engineering, The University of British Columbia\\
\texttt{yiliping@nbjl.nankai.edu.cn} \\
}

\begin{document}

\maketitle

\begin{abstract}

Model heterogeneous federated learning (MHeteroFL) enables FL clients to collaboratively train models with heterogeneous structures in a distributed fashion. However, existing MHeteroFL methods rely on training loss to transfer knowledge between the client model and the server model, resulting in limited knowledge exchange.
To address this limitation, we propose the \underline{Fed}erated model heterogeneous \underline{M}atryoshka \underline{R}epresentation \underline{L}earning (\methodname{}) approach for supervised learning tasks. 
It adds an auxiliary small \homo model shared by clients with \hetero local models. 
(1) The \gen and \pers \reps extracted by the two models' feature extractors are fused by a \pers lightweight \rep projector. This step enables \rep fusion to adapt to local data distribution.
(2) The fused \rep is then used to construct Matryoshka representations with multi-dimensional and multi-granular embedded \reps learned by the global \homo model header and the local \hetero model header. This step facilitates multi-perspective \rep learning and improves model learning capability.
Theoretical analysis shows that \methodname{} achieves a $\mathcal{O}(1/T)$ non-convex convergence rate.
Extensive experiments on benchmark datasets demonstrate its superior  
model accuracy with low communication and computational costs compared to seven state-of-the-art baselines. It achieves up to $8.48\%$ and $24.94\%$ accuracy improvement compared with the \sota and the best same-category baseline, respectively.
\end{abstract}

\section{Introduction}\label{sec:intro}
Traditional federated learning (FL) \cite{FedAvg} often relies on a central FL server to coordinate multiple data owners (a.k.a., FL clients) to train a global shared model without exposing local data. In each communication round, the server broadcasts the global model to the clients. A client trains it on its local data and sends the updated local model to the FL server. The server aggregates local models to produce a new global model. These steps are repeated until the global model converges. 

However, the above design cannot handle the following heterogeneity challenges \cite{FedGH} commonly found in practical FL applications: 
(1) Data heterogeneity \cite{PFL-yu,pFedLHNs,pFedKT,FedRRA,FFEDCL,SU-Net}: FL clients' local data often follow non-independent and identically distributions (non-IID). A single global model produced by aggregating local models trained on non-IID data might not perform well on all clients.
(2) System heterogeneity \cite{HeteroFL,FedPE,QSFL}: FL clients can have diverse system configurations in terms of computing power and network bandwidth. Training the same model structure among such clients means that the global model size must accommodate the weakest device, leading to sub-optimal performance on other more powerful clients.
(3) Model heterogeneity \cite{FedProto}: When FL clients are enterprises, they might have \hetero proprietary models which cannot be directly shared with others during FL training due to intellectual property (IP) protection concerns.

To address these challenges, the field of model heterogeneous federated learning (MHeteroFL) \citep{FedSSA,FedGH,FedLoRA,pFedES,pFedMoE,pFedAFM} has emerged. It enables FL clients to train local models with tailored structures suitable for local system resources and local data distributions. Existing MHeteroFL methods \citep{FML,FedKD} are limited in terms of knowledge transfer capabilities as they commonly leverage the training loss between server and client models for this purpose. This design leads to model performance bottlenecks, incurs high communication and computation costs, and risks exposing private local model structures and data.





\begin{figure}[ht]
\vspace{-0.5em}
\centering
\begin{minipage}[t]{0.5\linewidth}
\centering
\includegraphics[width=0.9\linewidth]{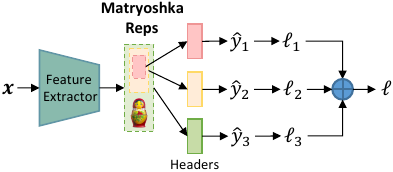}
\end{minipage}%
\begin{minipage}[t]{0.5\linewidth}
\centering
\includegraphics[width=\linewidth]{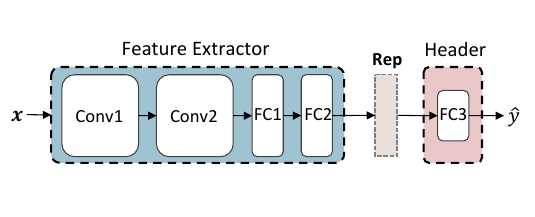}
\end{minipage}%
\vspace{-0.5em}
\caption{Left: Matryoshka Representation Learning. Right: Feature extractor and prediction header.}
\label{fig:split-MRL}
\vspace{-0.5em}
\end{figure}

Recently, Matryoshka Representation Learning (MRL) \cite{MRL} has emerged to tailor representation dimensions based on the computational and storage costs required by downstream tasks to achieve a near-optimal trade-off between model performance and inference costs. 
As shown in Figure~\ref{fig:split-MRL}(left), the \rep extracted by the feature extractor is constructed to form Matryoshka Representations involving a series of embedded representations ranging from low-to-high dimensions and coarse-to-fine granularities. Each of them is processed by a single output layer for calculating loss, and the sum of losses from all branches is used to update model parameters.
This design is inspired by the insight that people often first perceive the coarse aspect of a target before observing the details, with multi-perspective observations enhancing understanding.

Inspired by MRL, we address the aforementioned limitations of MHeteroFL by proposing the \underline{Fed}erated model heterogeneous \underline{M}atryoshka \underline{R}epresentation \underline{L}earning (\methodname{}) approach for supervised learning tasks. 
For each client, a shared global auxiliary \homo small model is added to interact with its \hetero local model. Both two models consist of a feature extractor and a prediction header, as depicted in Figure~\ref{fig:split-MRL}(right). \methodname{} has two key design innovations.
\textbf{(1) Adaptive Representation Fusion}: for each local data sample, the feature extractors of the two local models extract \gen and \pers representations, respectively. The two \reps are spliced and then mapped to a fused \rep by a lightweight \pers \rep projector adapting to local non-IID data.
\textbf{(2) Multi-Granularity Representation Learning}: the fused \rep is used to construct Matryoshka Representations involving multi-dimension and multi-granularity embedded representations, which are processed by the prediction headers of the two models, respectively. The sum of their losses is used to update all models, which enhances the model learning capability owing to multi-perspective \rep learning.

The \pers multi-granularity MRL enhances \rep knowledge interaction between the \homo global model and the \hetero client local model.
Each client's local model and data are not exposed during training for privacy-preservation. The server and clients only transmit the small \homo models, thereby incurring low communication costs. Each client only trains a small \homo model and a lightweight \rep projector in addition, incurring low extra computational costs. 
We theoretically derive the $\mathcal{O}(1/T)$ non-convex convergence rate of \methodname{} and verify that it can converge over time.
Experiments on benchmark datasets comparing \methodname{} against seven state-of-the-art baselines demonstrate its superiority. It improves model accuracy by up to $8.48\%$ and $24.94\%$ over the best baseline and the best same-category baseline, while incurring lower communication and computation costs.


\section{Related Work}\label{sec:related}
Existing MHeteroFL works can be divided into the following four categories.

\textbf{MHeteroFL with Adaptive Subnets.} These methods 
\citep{FedRolex,FLASH,InCo,HeteroFL,FjORD,Fed2,FedResCuE}
construct \hetero local subnets of the global model by parameter pruning or special designs to match with each client's local system resources. The server aggregates \hetero local subnets wise parameters to generate a new global model. In cases where clients hold black-box local models with heterogeneous structures not derived from a common global model, the server is unable to aggregate them.

\textbf{MHeteroFL with Knowledge Distillation.} These methods \citep{Cronus,FedGEMS,Fed-ET,FSFL,FCCL,DS-FL,FedMD,FedKT,FedDF,FedHeNN,FedKEM,KRR-KD,FedAUX,CFD,pFedHR,FedKEMF,KT-pFL}
often perform knowledge distillation on \hetero client models by leveraging a public dataset with the same data distribution as the learning task.
In practice, such a suitable public dataset can be hard to find.
Others \citep{FedIOD,FedGD,FedZKT,FedGen}
train a generator to synthesize a shared dataset to deal with this issue. However, this incurs high training costs. 
The rest ({\tt{FD}} \citep{FD}, {\tt{FedProto}} \citep{FedProto} and others \citep{HFD1,HFD2,FedGKT,FedGH,FedTGP}) share the intermediate information of client local data for knowledge fusion.

\textbf{MHeteroFL with Model Split.} These methods split models into feature extractors and predictors. Some \citep{FedMatch,FedRep,FedBABU,FedAlt/FedSim}
share \homo feature extractors across clients and personalize predictors, while others ({\tt{LG-FedAvg}} \citep{LG-FedAvg} and \citep{FedClassAvg,CHFL}) do the opposite. Such methods expose part of the local model structures, which might not be acceptable if the models are proprietary IPs of the clients.

\textbf{MHeteroFL with Mutual Learning.} These methods ({\tt{FedAPEN}} \citep{FedAPEN}, {\tt{FML}} \citep{FML}, {\tt{FedKD}} \citep{FedKD} and others \citep{FedME}) add a shared global \homo small model on top of each client's \hetero local model. For each local data sample, the distance of the outputs from these two models is used as the mutual loss to update model parameters. 
Nevertheless, the mutual loss only transfers limited knowledge between the two models, resulting in model performance bottlenecks.

The proposed \methodname{} approach further optimizes mutual learning-based MHeteroFL by enhancing the knowledge transfer between the server and client models. 
It achieves \pers adaptive \rep fusion and multi-perspective \rep learning, thereby facilitating more knowledge interaction across the two models and improving model performance.

\section{The Proposed \methodname{} Approach}
\methodname{} aims to tackle data, system, and model heterogeneity in supervised learning tasks, where a central FL server coordinates $N$ FL clients to train heterogeneous local models. The server maintains a global \homo small model $\mathcal{G}(\theta)$ shared by all clients. 
Figure~\ref{fig:FedMRL} depicts its workflow \footnote{Algorithm~\ref{alg:FedMRL} in Appendix~\ref{sec:alg} describes the \methodname{} algorithm.}:
\begin{circledenum}
    \item In each communication round, $K$ clients participate in FL (\emph{i.e.}, the client participant rate $C=K/N$).  The global \homo small model $\mathcal{G}(\theta)$ is broadcast to them. 
    \item Each client $k$ holds a \hetero local model $\mathcal{F}_k(\omega_k)$ ($\mathcal{F}_k(\cdot)$ is the \hetero model structure, and $\omega_k$ are \pers model parameters). Client $k$ simultaneously trains the \hetero local model and the global \homo small model on local non-IID data $D_k$ ($D_k$ follows the non-IID distribution $P_k$) via \pers Matryoshka Representations Learning with a \pers \rep projector $\mathcal{P}_k(\varphi_k)$. 
    \item The updated \homo small models are uploaded to the server for aggregation to produce a new global model for knowledge fusion across \hetero clients.
\end{circledenum}
The objective of \methodname{} is to minimize the sum of the loss from the combined models ($\mathcal{W}_k(w_k)=(\mathcal{G}(\theta)\circ\mathcal{F}_k(\omega_k)|\mathcal{P}_k(\varphi_k))$) on all clients, \emph{i.e.},
\begin{equation}\label{eq:objective}
\min _{\theta, \omega_{0, \ldots, N-1}} \sum_{k=0}^{N-1} \ell\left(\mathcal{W}_k\left(D_k ;\left(\theta \circ \omega_k \mid \varphi_k\right)\right)\right).
\end{equation}

These steps repeat until each client's model converges. After FL training, a client uses its local combined model without the global header for inference. \footnote{
Appendix~\ref{sec:model-inferene} provides experimental evidence for inference model selection.} 


\begin{figure}[t]
\centering
\includegraphics[width=\textwidth]{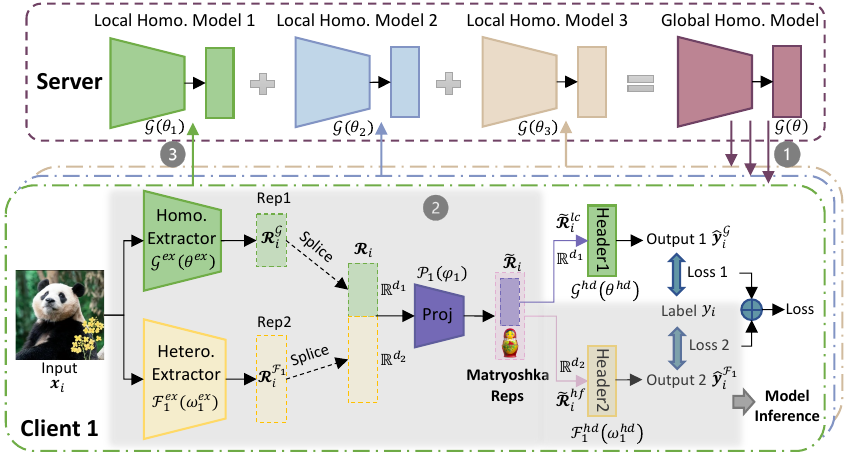}
\vspace{-0.5em}
\caption{The workflow of \methodname{}.}
\label{fig:FedMRL}
\vspace{-1em}
\end{figure}

\subsection{Adaptive Representation Fusion}
We denote client $k$'s \hetero local model feature extractor as $\mathcal{F}_k^{ex}(\omega_k^{ex})$, and prediction header as $\mathcal{F}_k^{hd}(\omega_k^{hd})$. We denote the \homo global model feature extractor as $\mathcal{G}^{ex}(\theta^{ex})$ and prediction header as $\mathcal{G}^{hd}(\theta^{hd})$. Client $k$'s local \pers \rep projector is denoted as $\mathcal{P}_k(\varphi_k)$.
In the $t$-th communication round, client $k$ inputs its local data sample $(\boldsymbol{x}_i,y_i)\in D_k$ into the two feature extractors to extract \gen and \pers \reps as:
\begin{equation}
    \boldsymbol{\mathcal{R}}_i^\mathcal{G}=\ \mathcal{G}^{ex}({\boldsymbol{x}_i;\theta}^{ex,t-1}),\boldsymbol{\mathcal{R}}_i^{\mathcal{F}_k}=\ \mathcal{F}_k^{ex}(\boldsymbol{x}_i;\omega_k^{ex,t-1}).
\end{equation}
The two extracted \reps $\boldsymbol{\mathcal{R}}_i^\mathcal{G} \in \mathbb{R}^{d_1}$ and $\boldsymbol{\mathcal{R}}_i^{\mathcal{F}_k} \in \mathbb{R}^{d_2}$ are spliced as:
\begin{equation}            \boldsymbol{\mathcal{R}}_i=\boldsymbol{\mathcal{R}}_i^\mathcal{G}\circ\boldsymbol{\mathcal{R}}_i^{\mathcal{F}_k}.
\end{equation}
Then, the spliced \rep is mapped into a fused \rep by the lightweight \rep projector $\mathcal{P}_k(\varphi_k^{t-1})$ as:
\begin{equation}
    {\widetilde{\boldsymbol{\mathcal{R}}}}_i=\mathcal{P}_k(\boldsymbol{\mathcal{R}}_i{;\varphi}_k^{t-1}),
\end{equation}
where the projector can be a one-layer linear model or multi-layer perceptron. The fused \rep ${\widetilde{\boldsymbol{\mathcal{R}}}}_i$ contains both \gen and \pers feature information. It has the same dimension as the client's local \hetero model \rep $\mathbb{R}^{d_2}$, which ensures the \rep dimension $\mathbb{R}^{d_2}$ and the client local \hetero model header parameter dimension $\mathbb{R}^{d_2\times L}$ ($L$ is the label dimension) match.

The \rep projector can be updated as the two models are being trained on local non-IID data. Hence, it achieves \pers \rep fusion adaptive to local data distributions.
Splicing the \reps extracted by two feature extractors can keep the relative semantic space positions of the \gen and \pers representations, benefiting the construction of multi-granularity Matryoshka Representations.
Owing to \rep splicing, the \rep dimensions of the two feature extractors can be different (\emph{i.e.}, $d_1 \leq d_2$). Therefore, we can vary the \rep dimension of the small \homo global model to improve the trade-off among model performance, storage requirement and communication costs.

In addition, each client's local model is treated as a black box by the FL server. When the server broadcasts the global \homo small model to the clients, each client can adjust the linear layer dimension of the \rep projector to align it with the dimension of the spliced representation. In this way, different clients may hold different \rep projectors. When a new model-agnostic client joins in \methodname{}, it can adjust its \rep projector structure for local model training. Therefore, \methodname{} can accommodate FL clients owning local models with diverse structures.

\subsection{Multi-Granular Representation Learning}
To construct multi-dimensional and multi-granular Matryoshka Representations, we further extract a low-dimension coarse-granularity \rep ${\widetilde{\boldsymbol{\mathcal{R}}}}_i^{lc}$ and a high-dimension fine-granularity \rep ${\widetilde{\boldsymbol{\mathcal{R}}}}_i^{hf}$ from the fused \rep ${\widetilde{\boldsymbol{\mathcal{R}}}}_i$. They align with the \rep dimensions $\{\mathbb{R}^{d_1}, \mathbb{R}^{d_2}\}$ of two feature extractors for matching the parameter dimensions $\{\mathbb{R}^{d_1\times L}, \mathbb{R}^{d_2\times L}\}$ of the two prediction headers,
\begin{equation}
    {\widetilde{\boldsymbol{\mathcal{R}}}}_i^{lc}={{\widetilde{\boldsymbol{\mathcal{R}}}}_i}^{1:d_1},{\widetilde{\boldsymbol{\mathcal{R}}}}_i^{hf}={{\widetilde{\boldsymbol{\mathcal{R}}}}_i}^{1:d_2}.
\end{equation}
The embedded low-dimension coarse-granularity \rep ${\widetilde{\boldsymbol{\mathcal{R}}}}_i^{lc} \in \mathbb{R}^{d_1}$ incorporates coarse \gen and \pers feature information. It is learned by the global \homo model header $\mathcal{G}^{hd}(\theta^{hd,t-1})$ (parameter space: $\mathbb{R}^{d_1\times L}$) with \gen prediction information to produce:
\begin{equation}
    {\hat{{y}}}_i^\mathcal{G}=\mathcal{G}^{hd}({\widetilde{\boldsymbol{\mathcal{R}}}}_i^{lc};\theta^{hd,t-1}).
\end{equation}
The embedded high-dimension fine-granularity \rep ${\widetilde{\boldsymbol{\mathcal{R}}}}_i^{hf} \in \mathbb{R}^{d_2}$ carries finer \gen and \pers feature information, which is further processed by the \hetero local model header $\mathcal{F}_k^{hd}(\omega_k^{hd,t-1})$ (parameter space: $\mathbb{R}^{d_2\times L}$) with \pers prediction information to generate:
\begin{equation}
{\hat{{y}}}_i^{\mathcal{F}_k}=\mathcal{F}_k^{hd}({\widetilde{\boldsymbol{\mathcal{R}}}}_i^{hf}; \omega_k^{hd,t-1}).
\end{equation}
We compute the losses $\ell$ (\emph{e.g.}, cross-entropy loss \citep{CEloss}) between the two outputs and the label $y_i$ as:
\begin{equation}
    \ell_i^\mathcal{G}=\ell({\hat{{y}}}_i^\mathcal{G},y_i),\ \ell_i^{\mathcal{F}_k}=\ell({\hat{{y}}}_i^{\mathcal{F}_k},y_i).
\end{equation}
Then, the losses of the two branches are weighted by their importance $m_i^\mathcal{G}$ and $m_i^{\mathcal{F}_k}$ and summed as:
\begin{equation}
\ell_i=m_i^\mathcal{G}\cdot\ell_i^\mathcal{G}+m_i^{\mathcal{F}_k}\cdot\ell_i^{\mathcal{F}_k}.
\end{equation}
We set $m_i^\mathcal{G}=m_i^{\mathcal{F}_k}=1$ by default to make the two models contribute equally to model performance.
The complete loss $\ell_i$ is used to simultaneously update the \homo global small model, the \hetero client local model, and the \rep projector via gradient descent:
\begin{equation}
    \begin{aligned}
        \theta_k^t&\gets\theta^{t-1}-\eta_\theta\nabla\ell_i, \\
\omega_k^t&\gets\omega_k^{t-1}-\eta_\omega\nabla\ell_i, \\
\varphi_k^t&\gets\varphi_k^{t-1}-\eta_\varphi\nabla\ell_i,
    \end{aligned}
\end{equation}
where $\eta_\theta,\eta_\omega,\ \eta_\varphi$ are the learning rates of the \homo global small model, the \hetero local model and the \rep projector. We set $\eta_\theta=\eta_\omega=\ \eta_\varphi$ by default to ensure stable model convergence.
In this way, the \gen and \pers fused \rep is learned from multiple perspectives, thereby improving model learning capability. 

\section{Convergence Analysis}
Based on notations, assumptions and proofs in Appendix~\ref{sec:theorey}, we analyse the convergence of \methodname{}.

\begin{lemma}
\label{lemma:localtraining}
    \textbf{Local Training.} Given Assumptions~\ref{assump:Lipschitz} and \ref{assump:Unbiased}, the loss of an arbitrary client's local model $w$ in local training round $(t+1)$ is bounded by:
    \begin{equation}
        \mathbb{E}[\mathcal{L}_{(t+1) E}] \leq \mathcal{L}_{t E+0}+(\frac{L_1 \eta^2}{2}-\eta) \sum_{e=0}^E\|\nabla \mathcal{L}_{t E+e}\|_2^2+\frac{L_1 E \eta^2 \sigma^2}{2}. 
    \end{equation}
\end{lemma}

\begin{lemma}
\label{lemma:aggregation}
\textbf{Model Aggregation.} Given Assumptions \ref{assump:Unbiased} and \ref{assump:BoundedVariation}, after local training round $(t+1)$, a client's loss before and after receiving the updated global \homo small models is bounded by:
\begin{equation}
 \mathbb{E}[\mathcal{L}_{(t+1)E+0}]\le\mathbb{E}[\mathcal{L}_{tE+1}]+{\eta\delta}^2.
\end{equation}
\end{lemma}

\begin{theorem}
\label{theorem:one-round}
\textbf{One Complete Round of FL.} 
Given the above lemmas, for any client, after receiving the updated global \homo small model, we have:
\begin{equation}
\mathbb{E}[\mathcal{L}_{(t+1) E+0}] \leq \mathcal{L}_{t E+0}+(\frac{L_1 \eta^2}{2}-\eta) \sum_{e=0}^E\|\nabla \mathcal{L}_{t E+e}\|_2^2+\frac{L_1 E \eta^2 \sigma^2}{2}+\eta \delta^2.
\end{equation}
\end{theorem}

\begin{theorem}
\label{theorem:non-convex}
\textbf{Non-convex Convergence Rate of FedMRL.} Given Theorem~\ref{theorem:one-round}, for any client and an arbitrary constant $\epsilon>0$, the following holds:
\begin{equation}
\begin{aligned}
\frac{1}{T} \sum_{t=0}^{T-1} \sum_{e=0}^{E-1}\|\nabla \mathcal{L}_{t E+e}\|_2^2 &\leq \frac{\frac{1}{T} \sum_{t=0}^{T-1}[\mathcal{L}_{t E+0}-\mathbb{E}[\mathcal{L}_{(t+1) E+0}]]+\frac{L_1 E \eta^2 \sigma^2}{2}+\eta \delta^2}{\eta-\frac{L_1 \eta^2}{2}}<\epsilon, \\
s.t. \   &\eta<\frac{2(\epsilon-\delta^2)}{L_1(\epsilon+E \sigma^2)} .
\end{aligned}
\end{equation}
\end{theorem}
Therefore, we conclude that any client's local model can converge at a non-convex rate of $\epsilon \sim \mathcal{O}(1/T)$ in \methodname{} if the learning rates of the \homo small model, the client local \hetero model and the \pers \rep projector satisfy the above conditions.

\section{Experimental Evaluation}\label{sec:exp}
We implement \methodname{} on Pytorch, and compare it with seven \sota MHeteroFL methods. The experiments are carried out over two benchmark supervised image classification datasets on $4$ NVIDIA GeForce 3090 GPUs (24GB Memory).\footnote{Codes are available in supplemental materials.} 

\subsection{Experiment Setup}\label{sec:setup}
\textbf{Datasets.} The benchmark datasets adopted are CIFAR-10 and CIFAR-100 \footnote{\scriptsize \url{https://www.cs.toronto.edu/\%7Ekriz/cifar.html}} \cite{cifar}, which are commonly used in FL image classification tasks for the evaluating existing MHeteroFL algorithms. CIFAR-10 has $60,000$ $32\times32$ colour images across $10$ classes, with $50,000$ for training and $10,000$ for testing. CIFAR-100 has $60,000$ $32\times32$ colour images across $100$ classes, with $50,000$ for training and $10,000$ for testing. We follow \cite{pFedHN} and \citep{FedAPEN} to construct two types of non-IID datasets. Each client's non-IID data are further divided into a training set and a testing set with a ratio of $8:2$.
\begin{itemize}
    \item \textbf{Non-IID (Class):} For CIFAR-10 with $10$ classes, we randomly assign $2$ classes to each FL client. For CIFAR-100 with $100$ classes, we randomly assign $10$ classes to each FL client. The fewer classes each client possesses, the higher the non-IIDness.
    \item \textbf{Non-IID (Dirichlet):} To produce more sophisticated non-IID data settings, for each class of CIFAR-10/CIFAR-100, we use a Dirichlet($\alpha$) function to adjust the ratio between the number of FL clients and the assigned data. A smaller $\alpha$ indicates more pronounced non-IIDness.
\end{itemize}

\textbf{Models.} We evaluate MHeteroFL algorithms under model-\homo and \hetero FL scenarios. \methodname{}'s \rep projector is a one-layer linear model (parameter space: $\mathbb{R}^{d2\times(d_1+d_2)}$).
\begin{itemize}
    \item \textbf{Model-Homogeneous FL:} All clients train CNN-1 in Table~\ref{tab:model-structures} (Appendix~\ref{sec:exp-models}). The \homo global small models in {\tt{FML}} and {\tt{FedKD}} are also CNN-1. The extra \homo global small model in \methodname{} is CNN-1 with a smaller \rep dimension $d_1$ (\emph{i.e.}, the penultimate linear layer dimension) than the CNN-1 model's \rep dimension $d_2$, $d_1 \leq d_2$.    
    \item \textbf{Model-Heterogeneous FL:} The $5$ \hetero models \{CNN-1, $\ldots$, CNN-5\} in Table~\ref{tab:model-structures} (Appendix~\ref{sec:exp-models}) are evenly distributed among FL clients. 
    The \homo global small models in {\tt{FML}} and {\tt{FedKD}} are the smallest CNN-5 models. The \homo global small model in \methodname{} is the smallest CNN-5 with a reduced \rep dimension $d_1$ compared with the CNN-5 model \rep dimension $d_2$, \emph{i.e.}, $d_1 \leq d_2$.
\end{itemize}

\textbf{Comparison Baselines.} We compare \methodname{} with \sota algorithms belonging to the following three categories of MHeteroFL methods:
\begin{itemize}
    \item {\tt{Standalone.}} Each client trains its \hetero local model only with its local data.
    \item \textbf{Knowledge Distillation Without Public Data:} {\tt{FD}} \cite{FD} and {\tt{FedProto}} \cite{FedProto}.
    \item \textbf{Model Split:} {\tt{LG-FedAvg}} \cite{LG-FedAvg}.
    \item \textbf{Mutual Learning:} {\tt{FML}} \cite{FML}, {\tt{FedKD}} \cite{FedKD} and {\tt{FedAPEN}} \citep{FedAPEN}. 
\end{itemize}

\textbf{Evaluation Metrics.} We evaluate MHeteroFL algorithms from the following three aspects:
\begin{itemize}
    \item \textbf{Model Accuracy.} We record the test accuracy of each client's model in each round, and compute the average test accuracy. 
    \item \textbf{Communication Cost.} We compute the number of parameters sent between the server and one client in one communication round, and record the required rounds for reaching the target average accuracy. The overall communication cost of one client for target average accuracy is the product between the cost per round and the number of rounds.
    \item \textbf{Computation Overhead.} We compute the computation FLOPs of one client in one communication round, and record the required communication rounds for reaching the target average accuracy. The overall computation overall for one client achieving the target average accuracy is the product between the FLOPs per round and the number of rounds.
\end{itemize}

\textbf{Training Strategy.} We search optimal FL hyperparameters and unique hyperparameters for all MHeteroFL algorithms. For FL hyperparameters, we test MHeteroFL algorithms with a $\{64, 128, 256, 512\}$ batch size, $\{1, 10\}$ epochs, $T=\{100, 500\}$ communication rounds and an SGD optimizer with a $0.01$ learning rate. The unique hyperparameter of \methodname{} is the \rep dimension $d_1$ of the \homo global small model, we vary $d_1=\{100,\ 150,...,500\}$ to obtain the best-performing \methodname{}.

\subsection{Results and Discussion}
We design three FL settings with different numbers of clients ($N$) and client participation rates ($C$): {($N=10, C=100\%$), ($N=50, C=20\%$), ($N=100, C=10\%$)} for both model-\homo and model-\hetero FL scenarios.

 
\subsubsection{Average Test Accuracy} 

Table~\ref{tab:compare-hetero} and Table~\ref{tab:compare-homo} (Appendix~\ref{sec:exp-homo}) show that \methodname{} consistently outperforms all baselines under both model-\hetero or \homo settings. It achieves up to a $8.48\%$ improvement in average test accuracy compared with the best baseline under each setting. Furthermore, it achieves up to a $24.94\%$ average test accuracy improvement than the best same-category (\emph{i.e.}, mutual learning-based MHeteroFL) baseline under each setting. These results demonstrate the superiority of \methodname{} in model performance owing to its adaptive \pers \rep fusion and multi-granularity \rep learning capabilities. 
Figure~\ref{fig:compare}(left six) shows that \methodname{} consistently achieves faster convergence speed and higher average test accuracy than the best baseline under each setting.

\subsubsection{Individual Client Test Accuracy}

\begin{figure}[t]
\centering
\begin{minipage}[t]{0.25\linewidth}
\centering
\includegraphics[width=1.45in]{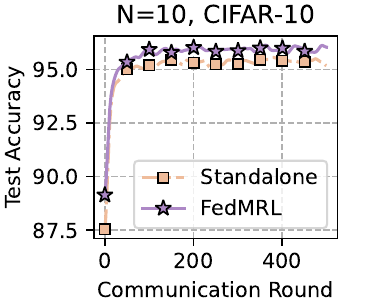}
\end{minipage}%
\begin{minipage}[t]{0.25\linewidth}
\centering
\includegraphics[width=1.45in]{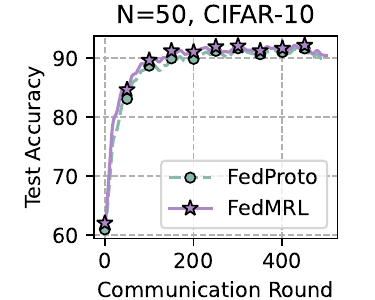}
\end{minipage}%
\begin{minipage}[t]{0.25\linewidth}
\centering
\includegraphics[width=1.45in]{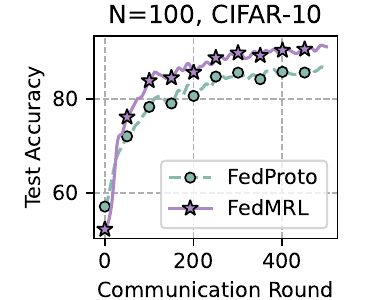}
\end{minipage}%
\begin{minipage}[t]{0.25\linewidth}
\centering
\includegraphics[width=1.45in]{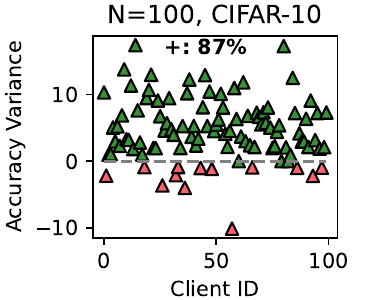}
\end{minipage}%

\begin{minipage}[t]{0.25\linewidth}
\centering
\includegraphics[width=1.45in]{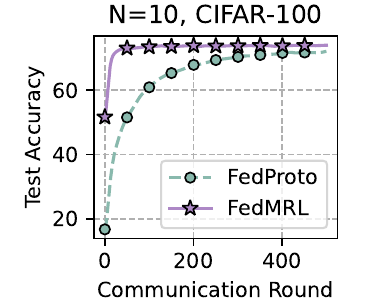}
\end{minipage}%
\begin{minipage}[t]{0.25\linewidth}
\centering
\includegraphics[width=1.45in]{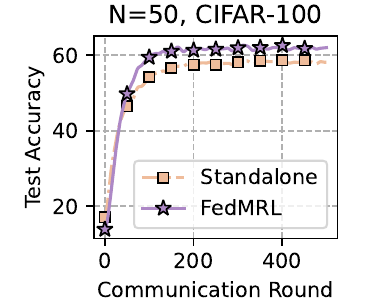}
\end{minipage}%
\begin{minipage}[t]{0.25\linewidth}
\centering
\includegraphics[width=1.45in]{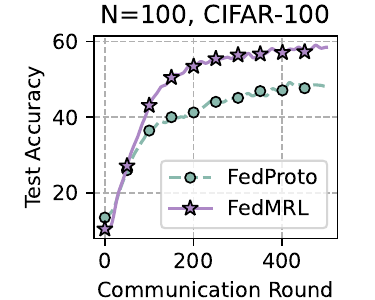}
\end{minipage}%
\begin{minipage}[t]{0.25\linewidth}
\centering
\includegraphics[width=1.45in]{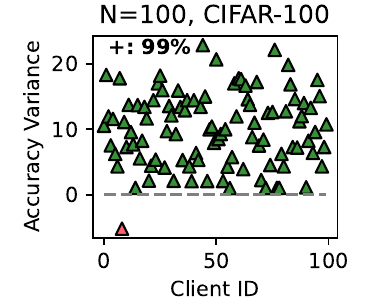}
\end{minipage}%
\vspace{-0.5em}
\caption{Left six: average test accuracy vs. communication rounds. Right two: individual clients' test accuracy (\%) differences (\methodname{} - {\tt{FedProto}}).}
\label{fig:compare}
\vspace{-1em}
\end{figure}

Figure~\ref{fig:compare}(right two) shows the difference between the test accuracy achieved by \methodname{} vs. the best-performing baseline {\tt{FedProto}} (\emph{i.e.}, \methodname{} - {\tt{FedProto}}) under ($N=100,C=10\%$) for each individual client. It can be observed that $87\%$ and $99\%$ of all clients achieve better performance under \methodname{} than under {\tt{FedProto}} on CIFAR-10 and CIFAR-100, respectively. This demonstrates that \methodname{} possesses stronger \persN capability than {\tt{FedProto}} owing to its adaptive \pers multi-granularity \rep learning design.

\begin{table}[t]
\centering
\caption{Average test accuracy (\%) in model-\hetero FL.}
\vspace{-0.5em}
\resizebox{\linewidth}{!}{%
\begin{tabular}{|l|cc|cc|cc|}
\hline
FL Setting                    & \multicolumn{2}{c|}{N=10, C=100\%}                                                                                                                   & \multicolumn{2}{c|}{N=50, C=20\%}                                                                                                                    & \multicolumn{2}{c|}{N=100, C=10\%}                                                                                                                   \\ \hline
Method                        & \multicolumn{1}{c|}{CIFAR-10}                                                      & CIFAR-100                                                     & \multicolumn{1}{c|}{CIFAR-10}                                                      & CIFAR-100                                                     & \multicolumn{1}{c|}{CIFAR-10}                                                      & CIFAR-100                                                     \\ \hline
Standalone                    & \multicolumn{1}{c|}{\cellcolor[HTML]{C0C0C0}{{96.53}}} & {72.53}                                  & \multicolumn{1}{c|}{{95.14}}                                  & \cellcolor[HTML]{C0C0C0}{{62.71}} & \multicolumn{1}{c|}{{91.97}}                                  & {53.04}                                  \\
LG-FedAvg~\citep{LG-FedAvg}                     & \multicolumn{1}{c|}{{96.30}}                                  & {72.20}                                  & \multicolumn{1}{c|}{{94.83}}                                  & {60.95}                                  & \multicolumn{1}{c|}{{91.27}}                                  & {45.83}                                  \\
FD~\citep{FD}                            & \multicolumn{1}{c|}{{96.21}}                                  & {-}                                      & \multicolumn{1}{c|}{{-}}                                      & {-}                                      & \multicolumn{1}{c|}{{-}}                                      & {-}                                      \\
FedProto~\citep{FedProto}                      & \multicolumn{1}{c|}{{96.51}}                                  & \cellcolor[HTML]{C0C0C0}{{72.59}} & \multicolumn{1}{c|}{\cellcolor[HTML]{C0C0C0}{{95.48}}} & {62.69}                                  & \multicolumn{1}{c|}{\cellcolor[HTML]{C0C0C0}{{92.49}}} & \cellcolor[HTML]{C0C0C0}{{53.67}} \\ \hline
FML~\citep{FML}                           & \multicolumn{1}{c|}{{30.48}}                                  & {16.84}                                  & \multicolumn{1}{c|}{{-}}                                      & {21.96}                                  & \multicolumn{1}{c|}{{-}}                                      & {15.21}                                  \\
FedKD~\citep{FedKD}                         & \multicolumn{1}{c|}{\cellcolor[HTML]{EFEFEF}{{80.20}}} & \cellcolor[HTML]{EFEFEF}{{53.23}} & \multicolumn{1}{c|}{\cellcolor[HTML]{EFEFEF}{{77.37}}} & \cellcolor[HTML]{EFEFEF}{{44.27}} & \multicolumn{1}{c|}{\cellcolor[HTML]{EFEFEF}{{73.21}}} & \cellcolor[HTML]{EFEFEF}{{37.21}} \\
FedAPEN~\citep{FedAPEN}                       & \multicolumn{1}{c|}{{-}}                             & {-}                             & \multicolumn{1}{c|}{{-}}                             & {-}                             & \multicolumn{1}{c|}{{-}}                                      & {-}                                      \\ \hline
\textbf{\methodname{}}               & \multicolumn{1}{c|}{\cellcolor[HTML]{9B9B9B}{\textbf{96.63}}} & \cellcolor[HTML]{9B9B9B}{\textbf{74.37}} & \multicolumn{1}{c|}{\cellcolor[HTML]{9B9B9B}{\textbf{95.70}}} & \cellcolor[HTML]{9B9B9B}{\textbf{66.04}} & \multicolumn{1}{c|}{\cellcolor[HTML]{9B9B9B}{\textbf{95.85}}} & \cellcolor[HTML]{9B9B9B}{\textbf{62.15}} \\ \hline
\textit{\methodname{}-Best B.}       & \multicolumn{1}{c|}{{\textit{0.10}}}  & {{\textit{1.78}}}                    & \multicolumn{1}{c|}{{\textit{0.22}}}  & {\textit{3.33}}                          & \multicolumn{1}{c|}{{\textit{3.36}}}  & {\underline{\textit{8.48}}}                          \\
\textit{\methodname{}-Best S.C.B.} & \multicolumn{1}{c|}{{\textit{16.43}}} & {{\textit{21.14}}}                   & \multicolumn{1}{c|}{{\textit{18.33}}} & {\textit{21.77}}                         & \multicolumn{1}{c|}{{\textit{22.64}}} & {\underline{\textit{24.94}}}                         \\ \hline
\end{tabular}
} \\
\raggedright 
\footnotesize ``-'': failing to converge. ``\mbox{
      \begin{tcolorbox}[colback=table_color2,
                  colframe=table_edge,
                  width=0.3cm,
                  height=0.25cm,
                  arc=0.25mm, auto outer arc,
                  boxrule=0.5pt,
                  left=0pt,right=0pt,top=0pt,bottom=0pt,
                 ]
      \end{tcolorbox}
      }'': the best MHeteroFL method.
``\mbox{
      \begin{tcolorbox}[colback=table_color3,
                  colframe=table_edge,
                  width=0.3cm,
                  height=0.25cm,
                  arc=0.25mm, auto outer arc,
                  boxrule=0.5pt,
                  left=0pt,right=0pt,top=0pt,bottom=0pt,
                 ]
      \end{tcolorbox}
      } Best B.'': the best baseline. ``\mbox{
      \begin{tcolorbox}[colback=table_color1,
                  colframe=table_edge,
                  width=0.3cm,
                  height=0.25cm,
                  arc=0.25mm, auto outer arc,
                  boxrule=0.5pt,
                  left=0pt,right=0pt,top=0pt,bottom=0pt,
                 ]
      \end{tcolorbox}
      } Best S.C.B.'': the best same-category (mutual learning-based MHeteroFL) baseline. The underscored values denote the largest accuracy improvement of \methodname{} across $6$ settings.

\label{tab:compare-hetero}
\vspace{-1em}
\end{table}

\begin{figure}[t]
\centering
\begin{minipage}[t]{0.3\linewidth}
\centering
\includegraphics[width=1.6in]{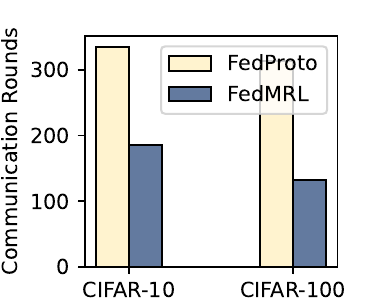}
\end{minipage}%
\begin{minipage}[t]{0.3\linewidth}
\centering
\includegraphics[width=1.6in]{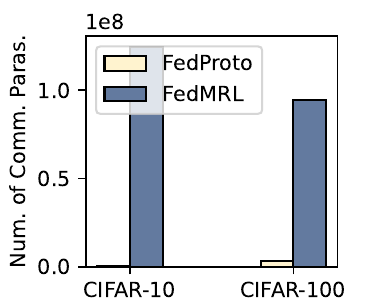}
\end{minipage}%
\begin{minipage}[t]{0.3\linewidth}
\centering
\includegraphics[width=1.6in]{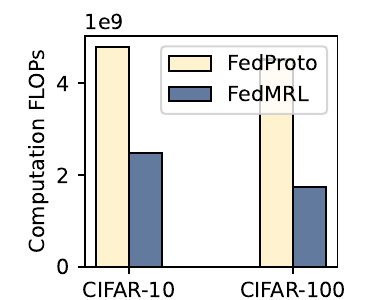}
\end{minipage}%
\vspace{-0.5em}
\caption{Communication rounds, number of communicated parameters, and computation FLOPs required to reach $90\%$ and $50\%$ average test accuracy targets on CIFAR-10 and CIFAR-100.}
\label{fig:comm-comp}
\vspace{-1em}
\end{figure}

\subsubsection{Communication Cost} 
We record the communication rounds and the number of parameters sent per client to achieve $90\%$ and $50\%$ target test average accuracy on CIFAR-10 and CIFAR-100, respectively. Figure~\ref{fig:comm-comp} (left) shows that \methodname{} requires fewer rounds and achieves faster convergence than {\tt{FedProto}}. Figure~\ref{fig:comm-comp} (middle) shows that \methodname{} incurs higher communication costs than {\tt{FedProto}} as it transmits the full \homo small model, while {\tt{FedProto}} only transmits each local seen-class average \rep between the server and the client. Nevertheless, \methodname{} with an optional smaller \rep dimension ($d_1$) of the \homo small model still achieves higher communication efficiency than same-category mutual learning-based MHeteroFL baselines ({\tt{FML}}, {\tt{FedKD}}, {\tt{FedAPEN}}) with a larger \rep dimension.

\subsubsection{Computation Overhead} 
We also calculate the computation FLOPs consumed per client to reach $90\%$ and $50\%$ target average test accuracy on CIFAR-10 and CIFAR-100, respectively. Figure~\ref{fig:comm-comp}(right) shows that \methodname{} incurs lower computation costs than {\tt{FedProto}}, owing to its faster convergence (\emph{i.e.}, fewer rounds) even with higher computation overhead per round due to the need to train an additional \homo small model and a linear \rep projector.

\subsection{Case Studies}
\subsubsection{Robustness to Non-IIDness (Class)}
\begin{figure}[t]
\centering
\begin{minipage}[t]{0.25\linewidth}
\centering
\includegraphics[width=1.45in]{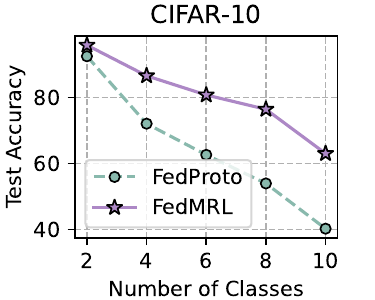}
\end{minipage}%
\begin{minipage}[t]{0.25\linewidth}
\centering
\includegraphics[width=1.45in]{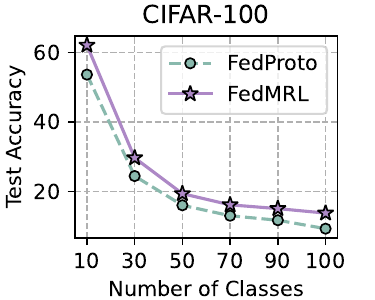}
\end{minipage}%
\begin{minipage}[t]{0.25\linewidth}
\centering
\includegraphics[width=1.45in]{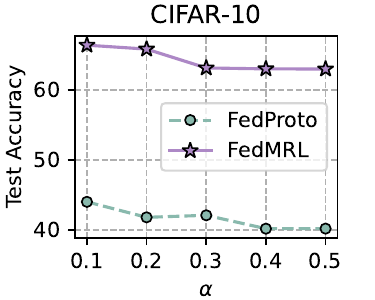}
\end{minipage}%
\begin{minipage}[t]{0.25\linewidth}
\centering
\includegraphics[width=1.45in]{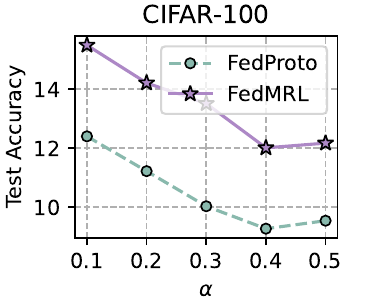}
\end{minipage}%
\vspace{-0.5em}
\caption{Robustness to non-IIDness (Class \& Dirichlet).}
\label{fig:case-noniid}
\vspace{-1em}
\end{figure}

We evaluate the robustness of \methodname{} to different non-IIDnesses as a result of the number of classes assigned to each client under the ($N=100, C=10\%$) setting. The fewer classes assigned to each client, the higher the non-IIDness.
For CIFAR-10, we assign $\{2, 4, \ldots, 10\}$ classes out of total $10$ classes to each client.
For CIFAR-100, we assign $\{10, 30, \ldots, 100\}$ classes out of total $100$ classes to each client.
Figure~\ref{fig:case-noniid}(left two) shows that \methodname{} consistently achieves higher average test accuracy than the best-performing baseline - {\tt{FedProto}} on both datasets, demonstrating its robustness to non-IIDness by class.

\begin{figure}[ht!]
\centering
\begin{minipage}[t]{0.25\linewidth}
\centering
\includegraphics[width=1.45in]{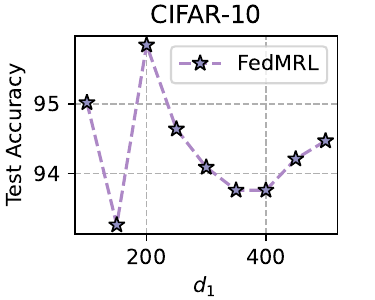}
\end{minipage}%
\begin{minipage}[t]{0.25\linewidth}
\centering
\includegraphics[width=1.45in]{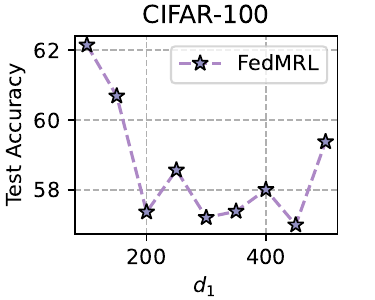}
\end{minipage}%
\begin{minipage}[t]{0.25\linewidth}
\centering
\includegraphics[width=1.45in]{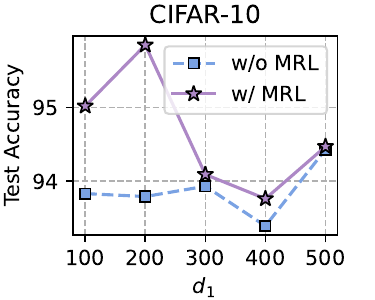}
\end{minipage}%
\begin{minipage}[t]{0.25\linewidth}
\centering
\includegraphics[width=1.45in]{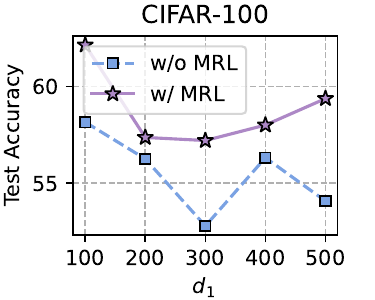}
\end{minipage}%
\vspace{-0.5em}
\caption{Left two: sensitivity analysis results. Right two: ablation study results.}
\label{fig:case-hyper}
\vspace{-1em}
\end{figure}

\subsubsection{Robustness to Non-IIDness (Dirichlet)}
We also test the robustness of \methodname{} to various non-IIDnesses controlled by $\alpha$ in the Dirichlet function under the ($N=100, C=10\%$) setting. A smaller $\alpha$ indicates a higher non-IIDness.
For both datasets, we vary $\alpha$ in the range of $\{0.1,\ldots,0.5\}$.
Figure~\ref{fig:case-noniid}(right two) shows that \methodname{} significantly outperforms {\tt{FedProto}} under all non-IIDness settings, validating its robustness to Dirichlet non-IIDness.

\subsubsection{Sensitivity Analysis - $d_1$}
\methodname{} relies on a  hyperparameter $d_1$ - the \rep dimension of the homogeneous small model.
To evaluate its sensitivity to $d_1$, we test \methodname{} with $d_1=\{100, 150, \ldots, 500\}$ under the ($N=100, C=10\%$) setting.
Figure~\ref{fig:case-hyper}(left two) shows that smaller $d_1$ values result in higher average test accuracy on both datasets. 
It is clear that a smaller $d_1$ also reduces communication and computation overheads, thereby helping \methodname{} achieve the best trade-off among model performance, communication efficiency, and computational efficiency.

\subsection{Ablation Study}
We conduct ablation experiments to validate the usefulness of MRL.
For \methodname{} with MRL, the global header and the local header learn multi-granularity representations. For \methodname{} without MRL, we directly input the \rep fused by the \rep projector into the client's local header for loss computation (\emph{i.e.}, we do not extract Matryoshka Representations and remove the global header).
Figure~\ref{fig:case-hyper}(right two) shows that \methodname{} with MRL consistently outperforms \methodname{} without MRL, demonstrating the effectiveness of the design to incorporate MRL into MHeteroFL. Besides, the accuracy gap between them decreases as $d_1$ rises. This shows that as the global and local headers learn increasingly overlapping \rep information, the benefits of MRL are reduced.

\section{Conclusions}
This paper proposes a novel MHeteroFL approach - \methodname{} - to jointly address data, system and model heterogeneity challenges in FL.
The key design insight is the addition of a global \homo small model shared by FL clients for enhanced knowledge interaction among \hetero local models. Adaptive \pers \rep fusion and multi-granularity Matryoshka Representations learning further boosts model learning capability.
The client and the server only need to exchange the \homo small model, while the clients' \hetero local models and data remain unexposed, thereby enhancing the preservation of both model and data privacy.
Theoretical analysis shows that \methodname{} is guaranteed to converge over time.
Extensive experiments demonstrate that \methodname{} significantly outperforms \sota models regarding test accuracy, while incurring low communication and computation costs. \footnote{
Appendix~\ref{sec:discuss} discusses \methodname{}'s privacy, communication and computation. 
Appendix~\ref{sec:lim} elaborates \methodname{}'s border impact and limitations.} 

{
\small

\bibliographystyle{plain}
\bibliography{ref.bib}

}

\newpage

\appendix

\section{Pseudo codes of \methodname{}}\label{sec:alg}
\vspace{-1em}
\begin{algorithm}[h!]
   \caption{\methodname{}}
\label{alg:FedMRL}
    {\bfseries Input:} $N$, total number of clients; $K$, number of selected clients in one round; $T$, total number of rounds; $\eta_\omega$, learning rate of client local heterogeneous models; $\eta_\theta$, learning rate of homogeneous small model; $\eta_\varphi$, learning rate of the representation projector.  \\
    {\bfseries Output:} client whole models removing the global header $[\mathcal{G}(\theta^{ex,T-1})\circ\mathcal{F}_0(\omega_0^{T-1})|\mathcal{P}_0(\varphi_0^{T-1}), \ldots, \mathcal{G}(\theta^{ex,T-1})\circ\mathcal{F}_{N-1}(\omega_{N-1}^{T-1})|\mathcal{P}_{N-1}(\varphi_{N-1}^{T-1})]$. \\
    
    \vspace{1em}
    Randomly initialize the global homogeneous small model $\mathcal{G}(\theta^\mathbf{0})$, client local heterogeneous models $[\mathcal{F}_0(\omega_0^0),\ldots,\mathcal{F}_{N-1}(\omega_{N-1}^0)]$ and local heterogeneous representation projectors $[\mathcal{P}_0(\varphi_0^0),\ldots,\mathcal{P}_{N-1}(\varphi_{N-1}^0)]$.

   \For{each round t=1,...,T-1}{
     // \textbf{Server Side}:\\
      $S^t$ $\gets$ Randomly sample $K$ clients from $N$ clients; \\
      Broadcast the global homogeneous small model $\theta^{t-1}$ to sampled $K$ clients; \\
      $\theta_k^t \gets$ \textbf{ClientUpdate}($\theta^{t-1}$); \\
         /* Aggregate Local Homogeneous Small Models */ \\
         $\theta^t=\sum_{k=0}^{K-1}{\frac{n_k}{n}\theta_k^t}$. \\
    \vspace{1em}
    // \textbf{ClientUpdate}:\\
    Receive the  global homogeneous small model $\theta^{t-1}$ from the server; \\
    \For{$k\in S^t$}{
      /* Local Training with MRL */  \\
     
     \For{$(\boldsymbol{x}_i,y_i)\in D_k$}{
        $\boldsymbol{\mathcal{R}}_i^\mathcal{G}=\ \mathcal{G}^{ex}({\boldsymbol{x}_i;\theta}^{ex,t-1}),\boldsymbol{\mathcal{R}}_i^{\mathcal{F}_k}=\ \mathcal{F}_k^{ex}(\boldsymbol{x}_i;\omega_k^{ex,t-1})$; \\
$\boldsymbol{\mathcal{R}}_i=\boldsymbol{\mathcal{R}}_i^\mathcal{G}\circ\boldsymbol{\mathcal{R}}_i^{\mathcal{F}_k}$;\\
${\widetilde{\boldsymbol{\mathcal{R}}}}_i=\mathcal{P}_k(\boldsymbol{\mathcal{R}}_i{;\varphi}_k^{t-1})$;\\
${\widetilde{\boldsymbol{\mathcal{R}}}}_i^{lc}={{\widetilde{\boldsymbol{\mathcal{R}}}}_i}^{1:d_1}, {\widetilde{\boldsymbol{\mathcal{R}}}}_i^{hf}={{\widetilde{\boldsymbol{\mathcal{R}}}}_i}^{1:d_2}$;\\
$ {\hat{{y}}}_i^\mathcal{G}=\mathcal{G}^{hd}({\widetilde{\boldsymbol{\mathcal{R}}}}_i^{lc};\theta^{hd,t-1});
 {\hat{{y}}}_i^{\mathcal{F}_k}=\mathcal{F}_k^{hd}(\omega_k^{hd,t-1})$;\\
 $\ell_i^\mathcal{G}=\ell({\hat{{y}}}_i^\mathcal{G},y_i); \ell_i^{\mathcal{F}_k}=\ell({\hat{{y}}}_i^{\mathcal{F}_k},y_i)$;\\
  $\ell_i=m_i^\mathcal{G}\cdot\ell_i^\mathcal{G}+m_i^{\mathcal{F}_k}\cdot\ell_i^{\mathcal{F}_k}$;\\
   $\theta_k^t\gets\theta^{t-1}-\eta_\theta\nabla\ell_i$; \\
$\omega_k^t\gets\omega_k^{t-1}-\eta_\omega\nabla\ell_i$; \\
$\varphi_k^t\gets\varphi_k^{t-1}-\eta_\varphi\nabla\ell_i$;\\
   }
     Upload updated local homogeneous small model $\theta_k^t$ to the server. \\
   }
}

\end{algorithm}
\vspace{-2em}

\section{Theoretical Proofs}\label{sec:theorey}
We first define the following additional notations.
$t \in \{0,\ldots,T-1\}$ denotes the $t$-th round. 
$e\in\{0,1,\ldots,E\}$ denotes the $e$-th iteration of local training.
$tE+0$ indicates that clients receive the global \homo small model $\mathcal{G}(\theta^{t})$ from the server before the $(t+1)$-th round's local training.
$tE+e$ denotes the $e$-th iteration of the $(t+1)$-th round's local training.
$tE+E$ marks the ending of the $(t+1)$-th round's local training. After that, clients upload their updated local \homo small models to the server for aggregation.
$\mathcal{W}_{k}(w_k)$ denotes the whole model trained on client $k$, including the global \homo small model $\mathcal{G}(\theta)$, the client $k$'s local \hetero model $\mathcal{F}_{k}(\omega_k)$, and the \pers \rep projector $\mathcal{P}_k(\varphi_k)$.
$\eta$ is the learning rate of the whole model trained on client $k$, including $\{\eta_\theta,\eta_\omega,\eta_{\boldsymbol{\varphi}}\}$.

\begin{assumption}\label{assump:Lipschitz}
\textbf{Lipschitz Smoothness}. The gradients of client $k$'s whole local model $w_k$ are $L1$--Lipschitz smooth \cite{FedProto},
\begin{equation}\label{eq:smmoth}
\begin{gathered}
\|\nabla \mathcal{L}_k^{t_1}(w_k^{t_1} ; \boldsymbol{x}, y)-\nabla \mathcal{L}_k^{t_2}(w_k^{t_2} ; \boldsymbol{x}, y)\| \leq L_1\|w_k^{t_1}-w_k^{t_2}\|, \\
\forall t_1, t_2>0, k \in\{0,1, \ldots, N-1\},(\boldsymbol{x}, y) \in D_k.
\end{gathered}
\end{equation}
The above formulation can be re-expressed as:
\begin{equation}
\mathcal{L}_k^{t_1}-\mathcal{L}_k^{t_2} \leq \langle\nabla \mathcal{L}_k^{t_2},(w_k^{t_1}-w_k^{t_2})\rangle+\frac{L_1}{2}\|w_k^{t_1}-w_k^{t_2}\|_2^2 .
\end{equation}
\end{assumption}

\begin{assumption} \label{assump:Unbiased}
\textbf{Unbiased Gradient and Bounded Variance}. Client $k$'s random gradient $g_{w,k}^t=\nabla \mathcal{L}_k^t(w_k^t; \mathcal{B}_k^t)$ ($\mathcal{B}$ is a batch of local data) is unbiased, 
\begin{equation}
\mathbb{E}_{\mathcal{B}_k^t \subseteq D_k}[g_{w,k}^t]=\nabla \mathcal{L}_k^t(w_k^t),
\end{equation}
and the variance of random gradient $g_{w,k}^t$ is bounded by:
\begin{equation}
\begin{split}
\mathbb{E}_{\mathcal{B}_k^t \subseteq D_k}[\|\nabla \mathcal{L}_k^t(w_k^t ; \mathcal{B}_k^t)-\nabla \mathcal{L}_k^t(w_k^t)\|_2^2] \leq \sigma^2.
\end{split}
\end{equation}    
\end{assumption}

\begin{assumption} \label{assump:BoundedVariation}
\textbf{Bounded Parameter Variation}. The parameter variations of the \homo small model $\theta_k^t$ and $\theta^t$ before and after aggregation at the FL server are bounded by:
\begin{equation}
     {\|\theta^t - \theta_k^{t}\|}_2^2 \leq \delta^2.
\end{equation}
\end{assumption} 



\subsection{Proof of Lemma~\ref{lemma:localtraining}}

\begin{proof}
    An arbitrary client $k$'s local whole model $w$ can be updated by $w_{t+1}=w_t-\eta g_{w,t}$ in the (t+1)-th round, and following Assumption~\ref{assump:Lipschitz}, we can obtain
\begin{equation}
\begin{aligned}
 \mathcal{L}_{t E+1} &\leq \mathcal{L}_{t E+0}+\langle\nabla \mathcal{L}_{t E+0},(w_{t E+1}-w_{t E+0})\rangle+\frac{L_1}{2}\|w_{t E+1}-w_{t E+0}\|_2^2 \\
& =\mathcal{L}_{t E+0}-\eta\langle\nabla \mathcal{L}_{t E+0}, g_{w, t E+0}\rangle+\frac{L_1 \eta^2}{2}\|g_{w, t E+0}\|_2^2 .   
\end{aligned}
\end{equation}

Taking the expectation of both sides of the inequality concerning the random variable $\xi_{tE+0}$, 
\begin{equation}
\begin{aligned}
 \mathbb{E}[\mathcal{L}_{t E+1}] &\leq \mathcal{L}_{t E+0}-\eta \mathbb{E}[\langle\nabla \mathcal{L}_{t E+0}, g_{w, t E+0}\rangle]+\frac{L_1 \eta^2}{2} \mathbb{E}[\|g_{w, t E+0}\|_2^2] \\
& \stackrel{(a)}{=} \mathcal{L}_{t E+0}-\eta\|\nabla \mathcal{L}_{t E+0}\|_2^2+\frac{L_1 \eta^2}{2} \mathbb{E}[\|g_{w, t E+0}\|_2^2] \\
& \stackrel{(b)}{\leq} \mathcal{L}_{t E+0}-\eta\|\nabla \mathcal{L}_{t E+0}\|_2^2+\frac{L_1 \eta^2}{2}(\mathbb{E}[\|g_{w, t E+0}\|]_2^2+\operatorname{Var}(g_{w, t E+0})) \\
& \stackrel{(c)}{=} \mathcal{L}_{t E+0}-\eta\|\nabla \mathcal{L}_{t E+0}\|_2^2+\frac{L_1 \eta^2}{2}(\|\nabla \mathcal{L}_{t E+0}\|_2^2+\operatorname{Var}(g_{w, t E+0})) \\
& \stackrel{(d)}{\leq} \mathcal{L}_{t E+0}-\eta\|\nabla \mathcal{L}_{t E+0}\|_2^2+\frac{L_1 \eta^2}{2}(\|\nabla \mathcal{L}_{t E+0}\|_2^2+\sigma^2) \\
& =\mathcal{L}_{t E+0}+(\frac{L_1 \eta^2}{2}-\eta)\|\nabla \mathcal{L}_{t E+0}\|_2^2+\frac{L_1 \eta^2 \sigma^2}{2}.
\end{aligned}
\end{equation}

(a), (c), (d) follow \assum \ref{assump:Unbiased} and (b) follows $Var(x)=\mathbb{E}[x^2]-(\mathbb{E}[x])^2$.

Taking the expectation of both sides of the inequality for the model $w$ over $E$ iterations, we obtain

\begin{equation}
\mathbb{E}[\mathcal{L}_{t E+1}] \leq \mathcal{L}_{t E+0}+(\frac{L_1 \eta^2}{2}-\eta) \sum_{e=1}^E\|\nabla \mathcal{L}_{t E+e}\|_2^2+\frac{L_1 E \eta^2 \sigma^2}{2} . 
\end{equation}
\end{proof}


\subsection{Proof of Lemma~\ref{lemma:aggregation}}

\begin{proof}
    \begin{equation}
\begin{aligned}
\mathcal{L}_{(t+1) E+0}& =\mathcal{L}_{(t+1) E}+\mathcal{L}_{(t+1) E+0}-\mathcal{L}_{(t+1) E} \\
& \stackrel{(a)}{\approx} \mathcal{L}_{(t+1) E}+\eta\|\theta_{(t+1) E+0}-\theta_{(t+1) E}\|_2^2 \\
& \stackrel{(b)}{\leq} \mathcal{L}_{(t+1) E}+\eta \delta^2.
\end{aligned}
\end{equation}

(a): we can use the gradient of parameter variations to approximate the loss variations, \emph{i.e.}, $\Delta\mathcal{L}\approx \eta\cdot \|\Delta \theta\|_2^2$. (b) follows \assum \ref{assump:BoundedVariation}.

Taking the expectation of both sides of the inequality to the random variable $\xi$, we obtain
\begin{equation}
    \mathbb{E}[\mathcal{L}_{(t+1)E+0}]\le\mathbb{E}[\mathcal{L}_{tE+1}]+{\eta\delta}^2.
\end{equation}

\end{proof}


\subsection{Proof of Theorem~\ref{theorem:one-round}}

\begin{proof}
    Substituting Lemma~\ref{lemma:localtraining} into the right side of Lemma~\ref{lemma:aggregation}'s inequality, we obtain
\begin{equation}\label{eq:theorem1}
\mathbb{E}[\mathcal{L}_{(t+1) E+0}] \leq \mathcal{L}_{t E+0}+(\frac{L_1 \eta^2}{2}-\eta) \sum_{e=0}^E\|\nabla \mathcal{L}_{t E+e}\|_2^2+\frac{L_1 E \eta^2 \sigma^2}{2}+\eta \delta^2.
\end{equation}
\end{proof}


\subsection{Proof of Theorem~\ref{theorem:non-convex}}

\begin{proof}
    Interchanging the left and right sides of Eq.~(\ref{eq:theorem1}), we obtain
\begin{equation}
\sum_{e=0}^E\|\nabla \mathcal{L}_{t E+e}\|_2^2 \leq \frac{\mathcal{L}_{t E+0}-\mathbb{E}[\mathcal{L}_{(t+1) E+0}]+\frac{L_1 E \eta^2 \sigma^2}{2}+\eta \delta^2}{\eta-\frac{L_1 \eta^2}{2}}.
\end{equation}

Taking the expectation of both sides of the inequality over rounds $t= [0, T-1]$ to $w$, we obtain
\begin{equation}
\frac{1}{T} \sum_{t=0}^{T-1} \sum_{e=0}^{E-1}\|\nabla \mathcal{L}_{t E+e}\|_2^2 \leq \frac{\frac{1}{T} \sum_{t=0}^{T-1}[\mathcal{L}_{t E+0}-\mathbb{E}[\mathcal{L}_{(t+1) E+0}]]+\frac{L_1 E \eta^2 \sigma^2}{2}+\eta \delta^2}{\eta-\frac{L_1 \eta^2}{2}}.
\end{equation}

Let $\Delta=\mathcal{L}_{t=0} - \mathcal{L}^* > 0$, then $\sum_{t=0}^{T-1}[\mathcal{L}_{t E+0}-\mathbb{E}[\mathcal{L}_{(t+1) E+0}]] \leq \Delta$, we can get 
\begin{equation}\label{eq:theorem2}
\frac{1}{T} \sum_{t=0}^{T-1} \sum_{e=0}^{E-1}\|\nabla \mathcal{L}_{t E+e}\|_2^2 \leq \frac{\frac{\Delta}{T}+\frac{L_1 E \eta^2 \sigma^2}{2}+\eta \delta^2}{\eta-\frac{L_1 \eta^2}{2}}.
\end{equation}

If the above equation converges to a constant $\epsilon$, \emph{i.e.},

\begin{equation}
\frac{\frac{\Delta}{T}+\frac{L_1 E \eta^2 \sigma^2}{2}+\eta \delta^2}{\eta-\frac{L_1 \eta^2}{2}}<\epsilon,
\end{equation}
then 
\begin{equation}
T>\frac{\Delta}{\epsilon(\eta-\frac{L_1 \eta^2}{2})-\frac{L_1 E \eta^2 \sigma^2}{2}-\eta \delta^2}.
\end{equation}

Since $T>0, \Delta>0$, we can get
\begin{equation}
\epsilon(\eta-\frac{L_1 \eta^2}{2})-\frac{L_1 E \eta^2 \sigma^2}{2}-\eta \delta^2>0.
\end{equation}

Solving the above inequality yields

\begin{equation}
\eta<\frac{2(\epsilon-\delta^2)}{L_1(\epsilon+E \sigma^2)}.
\end{equation}

Since $\epsilon,\ L_1,\ \sigma^2,\ \delta^2$ are all constants greater than 0, $\eta$ has solutions.
Therefore, when the learning rate $\eta =\{\eta_\theta,\eta_\omega,\eta_{\boldsymbol{\varphi}}\}$ satisfies the above condition, any client's local whole model can converge. Since all terms on the right side of Eq.~(\ref{eq:theorem2}) except for $1/T$ are constants, hence \methodname{}'s non-convex convergence rate is $\epsilon \sim \mathcal{O}(1/T)$.
\end{proof}

\section{More Experimental Details}
Here, we provide more experimental details of used model structures, more experimental results of model-\homo FL scenarios, and also the experimental evidence of inference model selection.

\subsection{Model Structures}\label{sec:exp-models}
Table~\ref{tab:model-structures} shows the structures of models used in experiments.
\begin{table}[!h]
\centering
\caption{Structures of $5$ heterogeneous CNN models.}
\resizebox{0.7\linewidth}{!}{%
\begin{tabular}{|l|c|c|c|c|c|}
\hline
Layer Name         & CNN-1    & CNN-2   & CNN-3   & CNN-4   & CNN-5   \\ \hline
Conv1              & 5$\times$5, 16   & 5$\times$5, 16  & 5$\times$5, 16  & 5$\times$5, 16  & 5$\times$5, 16  \\
Maxpool1              & 2$\times$2   & 2$\times$2  & 2$\times$2  & 2$\times$2  & 2$\times$2  \\
Conv2              & 5$\times$5, 32   & 5$\times$5, 16  & 5$\times$5, 32  & 5$\times$5, 32  & 5$\times$5, 32  \\
Maxpool2              & 2$\times$2   & 2$\times$2  & 2$\times$2  & 2$\times$2  & 2$\times$2  \\
FC1                & 2000     & 2000    & 1000    & 800     & 500     \\
FC2                & 500      & 500     & 500     & 500     & 500     \\
FC3                & 10/100   & 10/100  & 10/100  & 10/100  & 10/100  \\ \hline
model size & 10.00 MB & 6.92 MB & 5.04 MB & 3.81 MB & 2.55 MB \\ \hline
\end{tabular}%
}\\
\footnotesize Note: $5 \times 5$ denotes kernel size. $16$ or $32$ are filters in convolutional layers.
\label{tab:model-structures}
\vspace{-1em}
\end{table}

\subsection{Homogeneous FL Results}\label{sec:exp-homo}
Table~\ref{tab:compare-homo} presents the results of \methodname{} and baselines in model-\homo FL scenarios.
\begin{table}[!h]
\centering
\caption{Average test accuracy (\%) in model-\homo FL.}
\resizebox{\linewidth}{!}{%
\begin{tabular}{|l|cc|cc|cc|}
\hline
FL Setting                    & \multicolumn{2}{c|}{N=10, C=100\%}                                                                                                                   & \multicolumn{2}{c|}{N=50, C=20\%}                                                                                                                    & \multicolumn{2}{c|}{N=100, C=10\%}                                                                                                                   \\ \hline
Method                        & \multicolumn{1}{c|}{CIFAR-10}                                                      & CIFAR-100                                                     & \multicolumn{1}{c|}{CIFAR-10}                                                      & CIFAR-100                                                     & \multicolumn{1}{c|}{CIFAR-10}                                                      & CIFAR-100                                                     \\ \hline
Standalone                    & \multicolumn{1}{c|}{96.35}                                                         & \cellcolor[HTML]{C0C0C0}{74.32}                        & \multicolumn{1}{c|}{\cellcolor[HTML]{C0C0C0}{95.25}}                        & 62.38                                                         & \multicolumn{1}{c|}{\cellcolor[HTML]{C0C0C0}{92.58}}                        & \cellcolor[HTML]{C0C0C0}{54.93}                        \\
LG-FedAvg~\citep{LG-FedAvg}                     & \multicolumn{1}{c|}{\cellcolor[HTML]{C0C0C0}{96.47}}                        & 73.43                                                         & \multicolumn{1}{c|}{94.20}                                                         & 61.77                                                         & \multicolumn{1}{c|}{90.25}                                                         & 46.64                                                         \\
FD~\citep{FD}                            & \multicolumn{1}{c|}{96.30}                                                         & -                                                             & \multicolumn{1}{c|}{-}                                                             & -                                                             & \multicolumn{1}{c|}{-}                                                             & -                                                             \\
FedProto~\citep{FedProto}                      & \multicolumn{1}{c|}{95.83}                                                         & 72.79                                                         & \multicolumn{1}{c|}{95.10}                                                         & \cellcolor[HTML]{C0C0C0}{62.55}                        & \multicolumn{1}{c|}{91.19}                                                         & 54.01                                                         \\ \hline
FML~\citep{FML}                           & \multicolumn{1}{c|}{94.83}                                                         & 70.02                                                         & \multicolumn{1}{c|}{93.18}                                                         & 57.56                                                         & \multicolumn{1}{c|}{87.93}                                                         & 46.20                                                         \\
FedKD~\citep{FedKD}                         & \multicolumn{1}{c|}{94.77}                                                         & 70.04                                                         & \multicolumn{1}{c|}{92.93}                                                         & 57.56                                                         & \multicolumn{1}{c|}{\cellcolor[HTML]{EFEFEF}{90.23}}                        & \cellcolor[HTML]{EFEFEF}{50.99}                        \\
FedAPEN~\citep{FedAPEN}                       & \multicolumn{1}{c|}{\cellcolor[HTML]{EFEFEF}{95.38}}                        & \cellcolor[HTML]{EFEFEF}{71.48}                        & \multicolumn{1}{c|}{\cellcolor[HTML]{EFEFEF}{93.31}}                        & \cellcolor[HTML]{EFEFEF}{57.62}                        & \multicolumn{1}{c|}{87.97}                                                         & 46.85                                                         \\ \hline
\textbf{\methodname{}}               & \multicolumn{1}{c|}{\cellcolor[HTML]{9B9B9B}{\textbf{96.71}}} & \cellcolor[HTML]{9B9B9B}{\textbf{74.52}} & \multicolumn{1}{c|}{\cellcolor[HTML]{9B9B9B}{\textbf{95.76}}} & \cellcolor[HTML]{9B9B9B}{\textbf{66.46}} & \multicolumn{1}{c|}{\cellcolor[HTML]{9B9B9B}{\textbf{95.52}}} & \cellcolor[HTML]{9B9B9B}{\textbf{60.64}} \\ \hline
\textit{\methodname{}-Best B.}       & \multicolumn{1}{c|}{{\textit{0.24}}}  & { \textit{0.20}}                          & \multicolumn{1}{c|}{{\textit{0.51}}}  & {\textit{3.91}}                          & \multicolumn{1}{c|}{{\textit{2.94}}}  & {\underline{\textit{5.71}}}                         \\
\textit{\methodname{}-Best S.C.B.} & \multicolumn{1}{c|}{{\textit{1.33}}}  & {\textit{3.04}}                          & \multicolumn{1}{c|}{{\textit{2.45}}}  & {\textit{8.84}}                          & \multicolumn{1}{c|}{{\textit{5.29}}}  & {\underline{\textit{9.65}}}                          \\ \hline
\end{tabular}
} \\      
\raggedright 
\footnotesize ``-'': failing to converge. ``\mbox{
      \begin{tcolorbox}[colback=table_color2,
                  colframe=table_edge,
                  width=0.3cm,
                  height=0.25cm,
                  arc=0.25mm, auto outer arc,
                  boxrule=0.5pt,
                  left=0pt,right=0pt,top=0pt,bottom=0pt,
                 ]
      \end{tcolorbox}
      }'': the best MHeteroFL method.
``\mbox{
      \begin{tcolorbox}[colback=table_color3,
                  colframe=table_edge,
                  width=0.3cm,
                  height=0.25cm,
                  arc=0.25mm, auto outer arc,
                  boxrule=0.5pt,
                  left=0pt,right=0pt,top=0pt,bottom=0pt,
                 ]
      \end{tcolorbox}
      } Best B.'': the best baseline. ``\mbox{
      \begin{tcolorbox}[colback=table_color1,
                  colframe=table_edge,
                  width=0.3cm,
                  height=0.25cm,
                  arc=0.25mm, auto outer arc,
                  boxrule=0.5pt,
                  left=0pt,right=0pt,top=0pt,bottom=0pt,
                 ]
      \end{tcolorbox}
      } Best S.C.B.'': the best same-category (mutual learning-based MHeteroFL) baseline. The underscored values denote the largest accuracy improvement of \methodname{} across $6$ settings.
\label{tab:compare-homo}
\vspace{-1em}
\end{table}


\subsection{Inference Model Comparison}\label{sec:model-inferene}
\begin{figure}[!h]
\centering
\begin{minipage}[t]{0.5\linewidth}
\centering
\includegraphics[width=.75\linewidth]{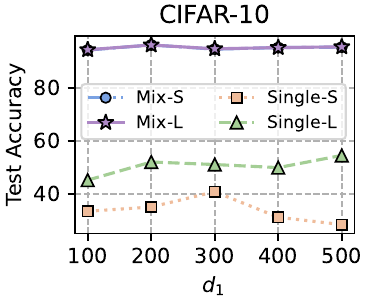}
\end{minipage}%
\begin{minipage}[t]{0.5\linewidth}
\centering
\includegraphics[width=.75\linewidth]{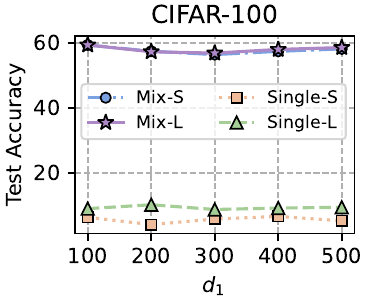}
\end{minipage}%
\vspace{-0.5em}
\caption{Accuracy of four optional inference models: mix-small (the whole model without the local header), mix-large (the whole model without the global header), single-small (the \homo small model), single-large (the client \hetero model).}
\label{fig:ablation-4models}
\vspace{-1em}
\end{figure}

There are $4$ alternative models for model inference in \methodname{}: (1) mix-small (the combination of the \homo small model, the client \hetero model's feature extractor, and the \rep projector, \emph{i.e.}, removing the local header), (2) mix-large (the combination of the \homo small model's feature extractor, the client \hetero model, and the \rep projector, \emph{i.e.}, removing the global header), (3) single-small (the \homo small model), (4) single-large (the client \hetero model). We compare their model performances under $(N=100, C=10\%)$ settings.
Figure~\ref{fig:ablation-4models} presents that mix-small has a similar accuracy to mix-large which is used as the default inference model, and they significantly outperform the single \homo small model and the single \hetero client model. Therefore, users can choose mix-small or mix-large for model inference based on their inference costs in practical applications.

\section{Discussion}\label{sec:discuss}
We discuss how \methodname{} tackles \heteroN and its privacy, communication and computation.

\textbf{Tackling Heterogeneity.} \methodname{} allows each client to tailor its \hetero local model according to its system resources, which addresses system and model heterogeneity. Each client achieves multi-granularity \rep learning adapting to local non-IID data distribution through a \pers \hetero \rep projector, alleviating data heterogeneity.

\textbf{Privacy.} The server and clients communicate the \homo small models while the \hetero local model is always stored in the client. Besides, \rep splicing enables the structures of the \homo global model and the \hetero local model to be not related. Therefore, the parameters and structure privacy of the \hetero client model is protected strongly. Meanwhile, the local data are always stored in clients for local training, so local data privacy is also protected.

\textbf{Communication Cost.} The server and clients transmit \homo small models with fewer parameters than the client's \hetero local model, consuming significantly lower communication costs in one communication round compared with transmitting complete local models like {\tt{FedAvg}}.

\textbf{Computational Overhead.} Except for training the client's \hetero local model, each client also trains the \homo global small model and a lightweight \rep projector which have far fewer parameters than the \hetero local model. The computational overhead in one training round is slightly increased. Since we design \pers Matryoshka Representations learning adapting to local data distribution from multiple perspectives, the model learning capability is improved, accelerating model convergence and consuming fewer training rounds. Therefore, the total computational cost may also be reduced.

\section{Broader Impacts and Limitations}\label{sec:lim}
\textbf{Broader Impacts.} \methodname{} improves model performance, communication and computational efficiency for heterogeneous federated learning while effectively protecting the privacy of the client \hetero local model and non-IID data. It can be applied in various practical FL applications.

\textbf{Limitations.} The multi-granularity embedded representations within Matryoshka Representations are processed by the global small model's header and the local client model's header, respectively. This increases the storage cost, communication costs and training overhead for the global header even though it only involves one linear layer.
In future work, we will follow the more effective Matryoshka Representation learning method (MRL-E) \citep{MRL}, removing the global header and only using the local model header to process multi-granularity Matryoshka Representations simultaneously, to enable a better trade-off among model performance and costs of storage, communication and computation.

\end{document}